\documentclass{article}

\PassOptionsToPackage{numbers, compress, sort}{natbib}

\usepackage[preprint]{neurips_2024}

\usepackage[utf8]{inputenc} 
\usepackage[T1]{fontenc}    
\usepackage{hyperref}       
\usepackage{url}            
\usepackage{booktabs}       
\usepackage{amsfonts}       
\usepackage{amssymb}
\usepackage{nicefrac}       
\usepackage{microtype}      
\usepackage{xcolor}         
\usepackage{graphicx}
\usepackage{amsmath}
\usepackage{mathtools}
\usepackage[capitalise]{cleveref}

\title{Universal Response \\ and Emergence of Induction in LLMs}

\author{%
  Niclas Luick \\
  University of Hamburg\\
  22761 Hamburg, Germany\\
  \texttt{nluick@physnet.uni-hamburg.de} \\
}

\begin{document}

\maketitle

\begin{abstract}

While induction is considered a key mechanism for in-context learning in LLMs, understanding its precise circuit decomposition beyond toy models remains elusive.
Here, we study the emergence of induction behavior within LLMs by probing their response to weak single-token perturbations of the residual stream. 
We find that LLMs exhibit a robust, universal regime in which their response remains scale-invariant under changes in perturbation strength, thereby allowing us to quantify the build-up of token correlations throughout the model. 
By applying our method, we observe signatures of induction behavior within the residual stream of \textit{Gemma-2-2B}, \textit{Llama-3.2-3B}, and \textit{GPT-2-XL}. 
Across all models, we find that these induction signatures gradually emerge within intermediate layers and identify the relevant model sections composing this behavior.
Our results provide insights into the collective interplay of components within LLMs and serve as a benchmark for large-scale circuit analysis.

\end{abstract}

\section{Introduction}

Over the last few years, the capabilities of large language models (LLMs) have dramatically improved - already reaching the level of human experts on a variety of tasks \cite{Learning11:online,reid2024gemini,dubey2024llama}.
In comparison to this remarkable rise of LLMs, our understanding of such models has remained relatively limited.

A promising approach to gain a better understanding of LLMs is mechanistic interpretability (MI), which aims at understanding a model's behavior through interpretable circuits of its components \cite{Mechanis63:online}.
While in toy models such circuits were found to explain a variety of model behavior including in-context learning \cite{wang2022interpretability,nanda2023progress,AMathema54:online,Incontex4:online}, understanding how such circuits can be extracted in larger models or for wider classes of behavior remains an open area of research \cite{conmy2023towards,lieberum2023does}.

What makes the composition of such circuits challenging for larger multi-layer models is a complex, non-linear interplay of multi-head attention (MHA), multi-layer perceptrons (MLPs), and skip-connections, via the residual stream \cite{vaswani2017attention}. 
This interplay is expected to give rise to superposition states of features \cite{elhage2022toy}, including attentional features \cite{Circuits50:online}, even across multiple layers \cite{ScalingM41:online}, and thereby significantly complicates such interpretability studies.
While recently, sparse autoencoders were successfully used to tackle the problem of superposition and find interpretable features as the variables for circuits in LLMs \cite{TowardsM13:online,ScalingM41:online,cunningham2023sparse,gao2024scaling}, many open questions remain \cite{Circuits98:online}.

Therefore, although we have a good understanding of individual model components \cite{bahdanau2014neural,lee2017interactive,clark2019does,liu2018visual,strobelt2018s,vig2019multiscale,rogers2021primer,gould2023successor,TowardsM13:online}, their collective downstream effect on the model behavior remains under active investigation \cite{jain2019attention,serrano2019attention,wiegreffe2019attention,abnar2020quantifying,rigotti2021attention,hao2021self,voita2019analyzing,chefer2021transformer,mcdougall2023copy,lad2024remarkable,ScalingM41:online}.
Consequently, while the field of MI is evolving rapidly and promising approaches to make progress on these questions are constantly emerging \cite{marks2024sparse,dunefsky2024transcoders,SparseCr31:online}, a full macroscopic understanding of LLMs is still lacking.

In this work, we examine the emergence of induction behavior, which is considered a key mechanism for in-context learning and therefore plays a  fundamental role for our understanding of LLMs \cite{Incontex4:online}.
Specifically, the induction mechanism enables models to correctly predict the next token in repeated sequences of the form $[A][B]\dots[A] \rightarrow [B]$, even for randomly chosen tokens $[A],[B]$.
In two-layer, attention-only toy models, this mechanism has been successfully reverse-engineered through a composition of \textit{previous token heads} followed by \textit{induction heads} \cite{AMathema54:online}.
So far, however, such circuit decompositions could not be scaled to larger, multi-layer transformer models, and the precise composition of induction behavior in LLMs remains under active investigation \cite{wang2022interpretability,mcdougall2023copy,ren2024identifying,singh2024needs,chen2024unveiling,crosbie2024induction,gould2023successor,Incontex4:online}.

\subsubsection*{Our contributions}

Here, we reveal the emergence of induction signatures within LLMs by probing the models' response to weak single-token perturbations of the residual stream. 
Specifically, we use repeated sequences of random tokens and find a strong response of the model for tokens preceding the perturbed token in the following sequence.
We observe such induction signatures within \textit{Gemma-2-2B}, \textit{Llama-3.2-3B}, and \textit{GPT-2-XL}, and across all models find a robust, universal regime in which the response remains scale-invariant under changes of perturbation strength.
We find that this scale-invariant regime extends throughout the residual stream, thereby allowing us to probe the composition of induction behavior within each model.
For all models, we observe the gradual emergence of induction signatures across intermediate layers and identify the relevant model sections composing this behavior.
Our results reveal qualitative differences in the composition of induction behavior in LLMs to guide future studies on large-scale circuit analysis \cite{SparseCr31:online}.

\section{Methods}

\subsection{Probing the response of LLMs}

\begin{figure}
  \centering
  \includegraphics[width=0.55\linewidth]{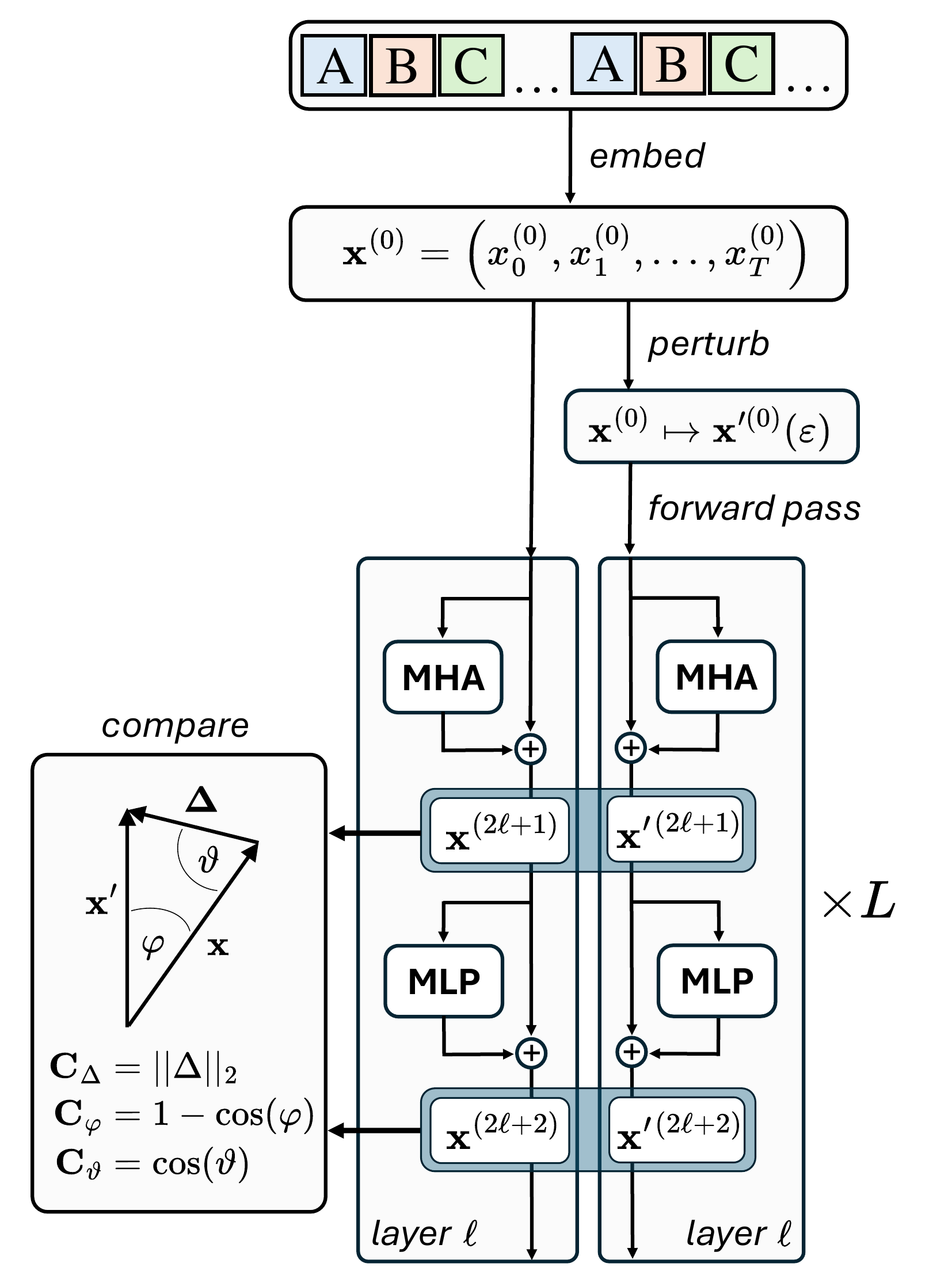}
  \caption{\textbf{Probing the response of LLMs.} We weakly perturb the residual stream vector $\mathbf{x}^{(0)}$ directly at the transformer input ($\ell=0$) for a single sequence position, $i=0,\dots, T-1$, by applying a scaling transformation $x_i{}^{(0)} \rightarrow {x}_i'{}^{(0)}(\varepsilon)=(1-\varepsilon)\cdot x_i{}^{(0)}$, while leaving all other sequence positions unchanged. To measure the response of the model, we compare the evolution of this perturbed vector $\mathbf{x'}{}^{(\ell)}$ to the unperturbed vector $\mathbf{x}^{(\ell)}$ for all downstream positions $\ell$ and token positions $j=0,\dots, T-1$ using three different response metrics $\mathbf{C}^{(\ell)}_{\Delta}$, $\mathbf{C}^{(\ell)}_{\varphi}$, $\mathbf{C}^{(\ell)}_{\vartheta} \in \mathbb{R}^{T\times T}$. As a benchmark, we apply this method to repeated subsequences of length $T_0 = T/2$, consisting of random tokens sampled uniformly from the vocabulary space.}
  \label{Fig1}
\end{figure}

Our method to probe the response of multi-layer transformer models is summarized in Fig. \ref{Fig1}.
Our scheme relies on perturbing the residual stream directly at the input of the transformer ($\ell=0$) and comparing the evolution of the perturbed vector $\mathbf{x}'^{(\ell)} \in \mathbb{R}^{T\times D}$ to the unperturbed state $\mathbf{x}^{(\ell)} \in \mathbb{R}^{T\times D}$ for each downstream position $\ell=1,2, \dots, 2L$, where $L$ is the number of layers, $T$ is the total sequence length, and $D$ is the model dimension.

As perturbation, we scale the residual stream vector, $x'^{(0)}_{i,j} (\varepsilon) = (x^{(0)}_0, \dots, (1-\varepsilon)\cdot x^{(0)}_i , \dots, x^{(0)}_T)_j$, at a single sequence position $i$, allowing us to continuously tune from weak $(\varepsilon \approx 0)$ to strong $(\varepsilon \approx 1)$ perturbations.
Due to the attention mechanism of transformers, this perturbation also influences the residual stream at other token positions, $j=0,\dots,T-1$, downstream in the model. 
We quantify the effect of perturbing at position $i$ onto tokens at position $j$ for each downstream location $\ell$ of the residual stream using the response matrices 
\begin{align}
(\mathbf{C}^{(\ell)}_{\Delta}(\varepsilon))_{i,j} &= ||\Delta_{i,j}^{(\ell)}(\varepsilon)||_2 \label{eq:CDelta} \,, \\
(\mathbf{C}_{\varphi}^{(\ell)}(\varepsilon))_{i,j} &=1-\cos(\varphi_{i,j}^{(\ell)}(\varepsilon)) \label{eq:Cphi} \,, \\
(\mathbf{C}^{(\ell)}_{\vartheta}(\varepsilon))_{i,j}&=\cos(\vartheta_{i,j}^{(\ell)}(\varepsilon)) \label{eq:Ctheta} \,,
\end{align}
which measure the $\ell^2$-norm of the difference $\Delta^{(\ell)}_{i,j}(\varepsilon) = x'^{(\ell)}_{i,j}(\varepsilon) - x^{(\ell)}_{,j} \in \mathbb{R}^{D}$, as well as the cosine similarities, $\cos(\varphi_{i,j}^{(\ell)}(\varepsilon))$ and $\cos(\vartheta_{i,j}^{(\ell)}(\varepsilon))$, computed for $x'^{(\ell)}_{i,j}(\varepsilon)$ and $\Delta^{(\ell)}_{i,j}(\varepsilon)$, respectively, relative to the unperturbed state $x^{(\ell)}_{,j}$.
Through this comparison to the unperturbed state, the response matrices (\ref{eq:CDelta}-\ref{eq:Ctheta}) quantify accumulated correlations between tokens, which contain the full downstream contribution of all model components up to the specific position $\ell$ in the residual stream.

\subsection{Probing induction behavior in \textit{Gemma-2-2B}}

To demonstrate that the introduced response matrices (\ref{eq:CDelta}-\ref{eq:Ctheta}) are indeed suitable metrics to study the emergence of induction behavior, we benchmark our method on repeated sequences of random tokens, using a pre-trained version of \textit{Gemma-2-2B} \cite{team2024gemma} as our model.
Figure \ref{Fig2}a and \ref{Fig2}d show typical response matrices in the weakly perturbed regime $(\varepsilon = \varepsilon_0 = 0.05)$, for a repeated subsequence of $T_0 = 64$ random tokens at the final position of the residual stream $(\ell = 2L)$.
We observe that both $\mathbf{C}_{\Delta}^{(2L)}$ and $\mathbf{C}_{\varphi}^{(2L)}$ display two distinct diagonal lines, offset by $\Delta j = 63$. 
This offset reveals strong correlations between tokens in the first subsequence and their corresponding predecessors at $\Delta j = T_0 - 1$ in the repeated subsequence, providing a clear signature of the model’s induction behavior \cite{AMathema54:online}.

To further quantify the model's response for specific position differences $\Delta j$, we calculate the response function
\begin{align}
\overline{C}_{\Delta, \varphi, \vartheta}^{(\ell)}(\varepsilon, \Delta j) = \sum_i(\mathbf{C}_{\Delta, \varphi, \vartheta}^{(\ell)}(\varepsilon))_{i,i+\Delta j} / (T-\Delta j)
\label{eq:CdeltaT}
\end{align}
for each response matrix (\ref{eq:CDelta}-\ref{eq:Ctheta}) by averaging along diagonals offset by a position difference $\Delta j = 0, \dots, T-1$ from the position of the perturbation.
In line with our previous findings, we observe a pronounced peak in both $\overline{C}{}^{(2L)}_{\Delta}(\varepsilon, \Delta j)$ and $\overline{C}{}^{(2L)}_{\varphi}(\varepsilon, \Delta j)$ at $\Delta j = 63$ (s. Fig. \ref{Fig2}b, \ref{Fig2}e).
This peak is asymmetric, characterized by a sharp initial rise followed by a more gradual decay as $\Delta j$ increases, similar to the decay following the initial peak at $\Delta j = 0$.

\section{Scale-invariant response}

\begin{figure}[t]
  \centering
  \includegraphics[width=\linewidth]{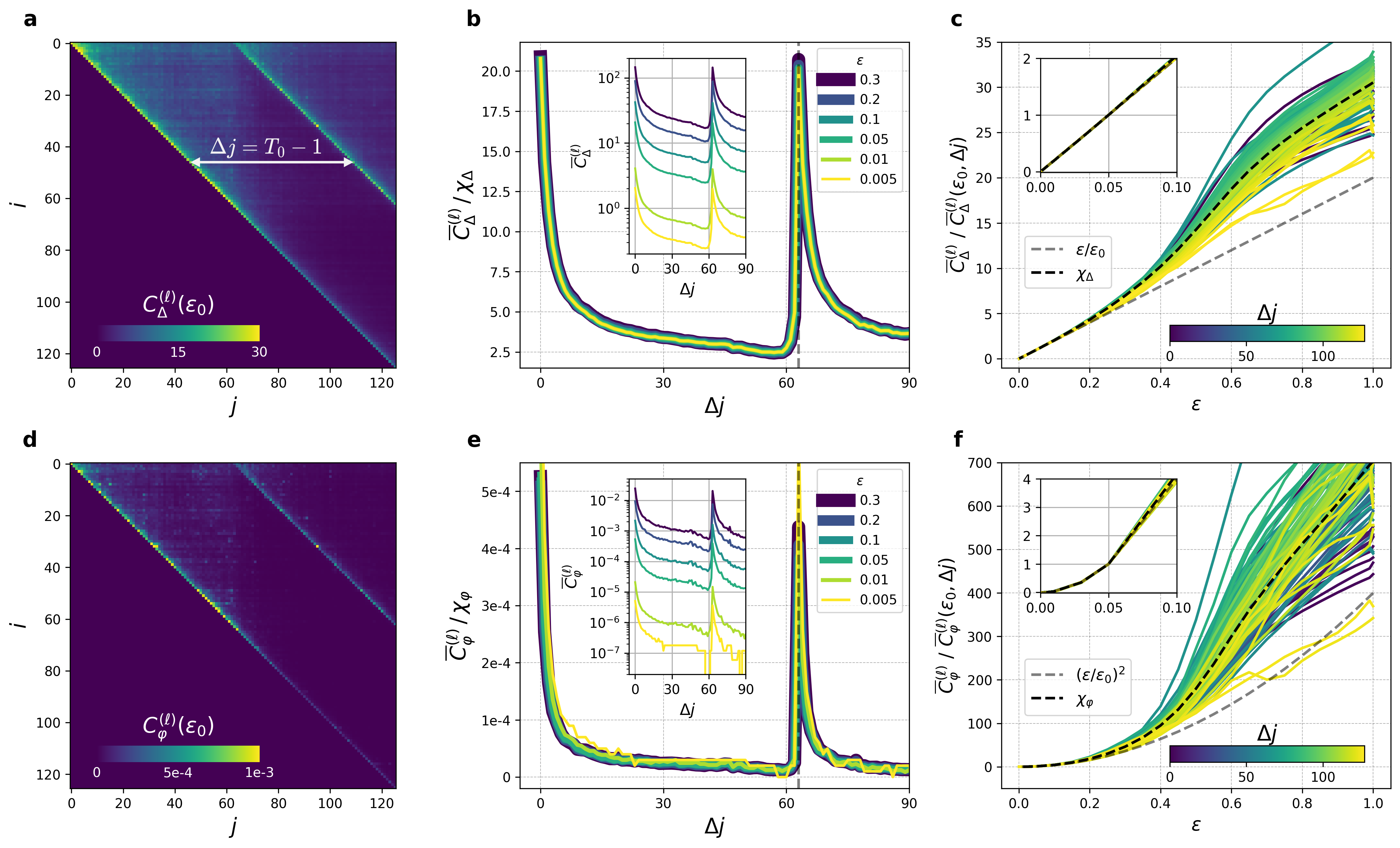}
  \caption{\textbf{Scale-invariant response.} 
  \textbf{(a, d)} Response matrices $\mathbf{C}^{(\ell)}_{\Delta}$ (a) and $\mathbf{C}^{(\ell)}_{\varphi}$ (d) for a repeated sequence of $T_0 = 64$ random tokens, a weak perturbation $(\varepsilon=\varepsilon_0=0.05)$, and the last position of the residual stream $(\ell=2L)$, using a pre-trained model of \textit{Gemma-2-2B} ($L=26$). \textbf{(b, e)} The rescaled response functions $\overline{C}{}^{(\ell)}_{\Delta} / \chi_{\Delta}$ (b) and $\overline{C}{}^{(\ell)}_{\varphi} / \chi_{\varphi}$ (e) collapse the unscaled data (insets) onto a single, $s$-independent curve with a pronounced peak at $\Delta j = T_0-1$ (dashed vertical lines) as a signature of induction behavior. \textbf{(c, f)} For small perturbations $(\varepsilon < 0.1)$, the ratios $\overline{C}{}^{(\ell)}_{\Delta} / \overline{C}{}^{(\ell)}_{\Delta}(\varepsilon_0, \Delta j)$ (c) and $\overline{C}{}^{(\ell)}_{\varphi} / \overline{C}{}^{(\ell)}_{\varphi}(\varepsilon_0, \Delta j)$ (f) are independent of $\Delta j$ and well-approximated by $\chi_{\Delta} \approx \varepsilon/\varepsilon_0$ (c, grey dashed line) and $\chi_{\Delta} \approx \left(\varepsilon/\varepsilon_0\right)^2$ (f, grey dashed line), thereby demonstrating the scale-invariance of the response. All data shown in (a-f) is obtained by averaging over a batch of $32$ sequences.}
  \label{Fig2}
\end{figure}

To test the robustness of our introduced response metrics, we repeat the experimental protocol described above and measure the response functions in the entire range from weak to strong perturbations (s. Fig. \ref{Fig2}b, \ref{Fig2}e).
Strikingly, we find that the shape of $\overline{C}{}^{(2L)}_{\Delta}$ and $\overline{C}{}^{(2L)}_{\varphi}$ remains unchanged over a wide range in perturbation strength $(0.005 \le \varepsilon \le 0.3)$, corresponding to a variation of two to three orders of magnitude in the amplitude of the response.
This result confirms that we can use the response functions as a robust, quantitative probe of induction behavior within our model.

We further investigate the scaling behavior of both response functions by calculating the ratios $\overline{C}{}^{(2L)}_{\Delta} / \overline{C}{}^{(2L)}_{\Delta}(\varepsilon_0, \Delta j)$ and $\overline{C}{}^{(2L)}_{\varphi} / \overline{C}{}^{(2L)}_{\varphi}(\varepsilon_0, \Delta j)$, where $\varepsilon_0 = 0.05$ is chosen as a reference value for a weak perturbation (s. Fig. \ref{Fig2}c, \ref{Fig2}f).
For sufficiently weak perturbation strengths $(\varepsilon < 0.1)$, we find that these ratios each collapse onto a single curve, only determined by the value of $\varepsilon$, and nearly independent of $\Delta j$.
In line with our previous findings, this result confirms that our response functions are simply rescaled in amplitude under changes of the scale parameter $\varepsilon$.
We determine the corresponding scaling functions $\chi_{\Delta}$ and $\chi_{\varphi}$ by averaging the response ratios over all values of $\Delta j$, and find that these functions are well-approximated by $\chi_{\Delta} \approx \varepsilon/\varepsilon_0$ and $\chi_{\varphi} \approx \left(\varepsilon/\varepsilon_0\right)^2$ for sufficiently weak perturbations $(\varepsilon<0.1)$.
Therefore, our response functions are in fact scale-invariant in this weak perturbation regime \cite{STANLEY200060}, i.e.,
\begin{align}
\overline{C}^{(\ell)}_{\Delta}(\lambda \varepsilon, \Delta j) &= \lambda \cdot \overline{C}^{(\ell)}_{\Delta}(\varepsilon, \Delta j), 
\label{eq:scale1}
\\
\overline{C}^{(\ell)}_{\varphi}(\lambda \varepsilon, \Delta j) &= \lambda^2 \cdot \overline{C}^{(\ell)}_{\varphi}(\varepsilon, \Delta j).
\label{eq:scale2}
\end{align}
From the observed scaling behavior of the response, we further conclude that the perturbed state at the final position of the model $(\ell = 2L)$ can be approximated as 
\begin{equation}
\mathbf{x}'^{(\ell)}(\varepsilon) \approx \mathbf{x}^{(\ell)} + \varepsilon \cdot \mathbf{\Delta}_0^{(\ell)}(\varepsilon), \quad \text{for } \varepsilon \leq 0.1,
\label{eq:xprime}
\end{equation}
using a normalized response vector $\mathbf{\Delta}_0^{(\ell)}(\varepsilon) \in \mathbb{R}^{T\times D}$, which is nearly orthogonal to $\mathbf{x}^{(\ell)}$, and with $\varepsilon$-independent amplitude, for all values of $\Delta j$.
This result suggests that the induction mechanism of our model is encoded in an orthogonal subspace of the residual stream. 

\begin{figure}
  \centering
  \includegraphics[width=\linewidth]{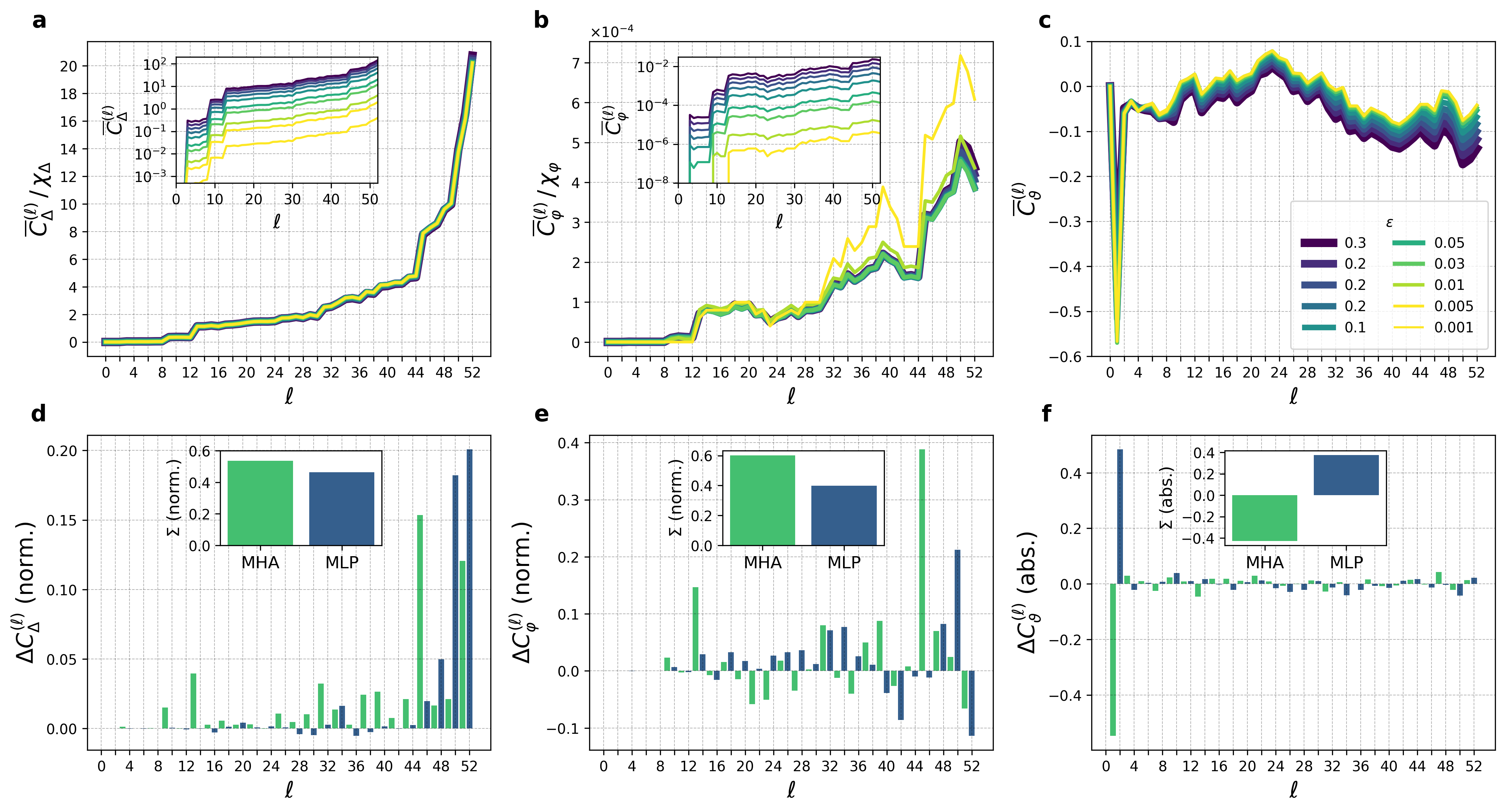}
  \caption{\textbf{Scale-invariant response across model layers.} \textbf{(a, b, c)} Evolution of the response functions $\overline{C}{}^{(\ell)}_{\Delta}$ (a, inset), $\overline{C}{}^{(\ell)}_{\varphi}$ (b, inset), and $\overline{C}{}^{(\ell)}_{\vartheta}$ (c) across the residual stream of \textit{Gemma-2-2B}, for various values of the scale parameter $\varepsilon$, at a fixed value of $\Delta j = T_0 - 1$, using the same sequence data as in Fig. \ref{Fig2}. Rescaling with $\ell$-independent scaling functions $\chi_{\Delta}$ and $\chi{_\varphi}$, collapses $\overline{C}{}^{(\ell)}_{\Delta}$ (a) and $\overline{C}{}^{(\ell)}_{\varphi}$ (b) onto a single curve, nearly independent of the value of $\varepsilon$ over a wide range in perturbation strength. \textbf{(d, e, f)} Increments $\Delta C^{(\ell)}_{\Delta, \varphi, \vartheta} = \overline{C}{}^{(\ell)}_{\Delta, \varphi, \vartheta} - \overline{C}{}^{(\ell-1)}_{\Delta, \varphi, \vartheta}$ reveal the contribution of individual MHA (green bars) and MLP sublayers (blue bars) to the response functions $\overline{C}{}^{(\ell)}_{\Delta}$ (d), $\overline{C}{}^{(\ell)}_{\varphi}$ (e), and $\overline{C}{}^{(\ell)}_{\vartheta}$ (f), for each position $\ell$ in the model. The values of $\Delta C^{(\ell)}_{\Delta}$ and $\Delta C^{(\ell)}_{\varphi}$ are each normalized such that $\sum_{\ell}\Delta C_{\Delta, \varphi}^{(\ell)} = 1$. Partial sums over increments, $\Sigma = \sum_{\ell \text{ even/odd}}\Delta C_{\Delta, \varphi, \vartheta}^{(\ell)}$, reveal the total contribution of MHA ($\ell$ odd) and MLP sublayers ($\ell$ even) to the response (insets). All data shown is obtained by averaging over a batch of $32$ sequences.}
  \label{Fig3}
\end{figure}

To test if the scale-invariance of the response also holds within intermediate layers, we measure $\overline{C}{}_{\Delta}^{(\ell)}$ and $\overline{C}{}_{\varphi}^{(\ell)}$ for all positions $\ell$ within the residual stream, at a fixed value of $\Delta j = T_0-1$, and rescale the data with the previously determined scaling functions $\chi_{\Delta}$ and $\chi_{\varphi}$ (s. Fig. \ref{Fig3}).
Indeed, for all values of $\ell$, we find that the rescaled response functions $\overline{C}{}^{(\ell)}_{\Delta}/ \chi_{\Delta}$ and $\overline{C}{}^{(\ell)}_{\varphi}/ \chi_{\varphi}$ remain nearly independent of $\varepsilon$ over a wide range in perturbation strength.
Therefore, the scaling functions $\chi_{\Delta}$ and $\chi_{\varphi}$ are in fact universal across all layers of the model, and the scale-invariance of the response, and hence the validity of Eqs. (\ref{eq:scale1}-\ref{eq:xprime}), holds for all values of $\ell = 1,\dots, 2L$. 

These results demonstrate the robustness of our method in quantifying the composition of model behavior across the residual stream.
Specifically, we find that our induction signals in $\overline{C}{}^{(\ell)}_{\Delta}$ and $\overline{C}{}^{(\ell)}_{\varphi}$ both gradually increase throughout the model - with noticeable jumps in both metrics in layer 6 ($\ell=13$) and layer 22 ($\ell=45$) due to MHA outputs, and another sharp increase in $\overline{C}{}^{(\ell)}_{\Delta}$ in the final two layers, predominately due to MLP outputs.
Accumulated over the entire residual stream, both MHA and MLP sublayers contribute nearly equally to the observed induction signal at $\Delta j = T_0 - 1$.

Finally, we conclude our analysis of scale-invariance, by measuring the response function $\overline{C}{}^{(\ell)}_{\vartheta}$ which quantifies the alignment of the response vector $\mathbf{\Delta}_0^{(\ell)}$ compared to the unperturbed state $\mathbf{x}^{(\ell)}$, using the same experimental parameters as above (s. Fig. \ref{Fig3}c, \ref{Fig3}f).
Across the entire residual stream, we observe that the response converges to $\overline{C}{}^{(\ell)}_{\vartheta}(\varepsilon_0, \Delta j)$ for small perturbations, and is only determined by the values of $\ell$, and $\Delta j$.
Furthermore, apart from the first layer of the model, we find that $|\overline{C}{}^{(\ell)}_{\vartheta}(\varepsilon_0, \Delta j)|< 0.1 $, thereby confirming that the response vector $\mathbf{\Delta}_0^{(\ell)}$ indeed remains nearly orthogonal to the unperturbed state $\mathbf{x}^{(\ell)}$ throughout the model.

\begin{figure}
  \centering
  \includegraphics[width=\linewidth]{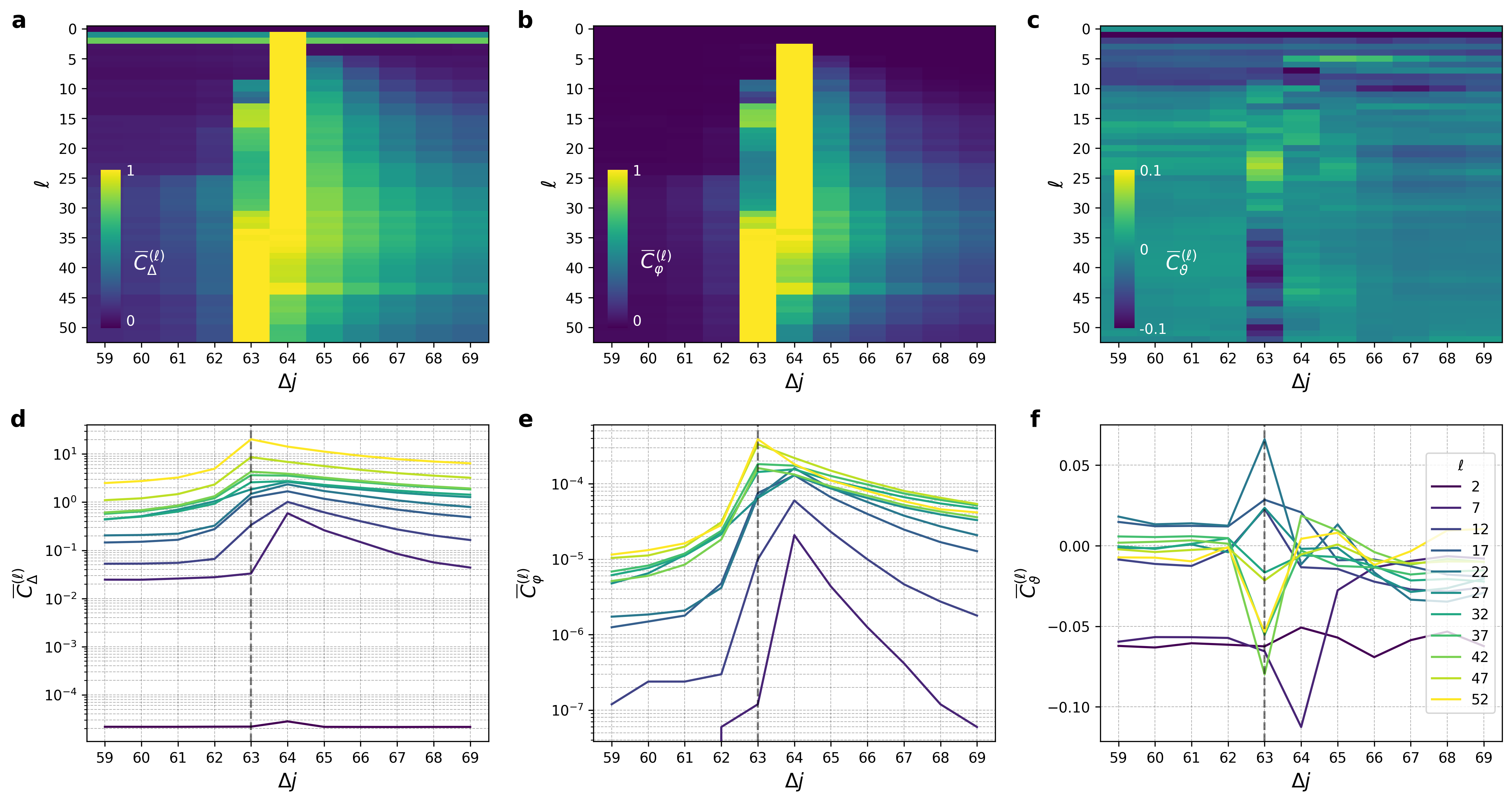}
  \caption{\textbf{Emergence of induction signatures in \textit{Gemma-2-2B}.} \textbf{(a-f)} Response functions $\overline{C}{}^{(\ell)}_{\Delta}$ (a, d), $\overline{C}{}^{(\ell)}_{\varphi}$ (b, e), and $\overline{C}{}^{(\ell)}_{\vartheta}$ (c, f), for varying values of $\Delta j$ and $\ell$, using a weak perturbation $(\varepsilon=0.05)$, and a repeated sequence of $T_0=64$ random tokens. Each row of images shown in (a, b) is normalized to the maximum value of $\overline{C}{}^{(\ell)}_{\Delta, \varphi}$ within the range of $\Delta j = 59 - 69$. As a signature for the onset of induction behavior within the model around $\ell\gtrsim 30$, we observe a shift from correlations between the same tokens $(\Delta j = T_0)$ to their previous tokens $(\Delta j = T_0 - 1)$ of the repeated subsequence. All data shown is obtained by averaging over a batch of $32$ sequences.}
  \label{Fig4}
\end{figure}

\section{Emergence of induction signatures}

Now that we have established the robustness of our method, we finally apply it to qualitatively show the emergence of induction signatures within LLMs.
To do this, we map out the evolution of the model's response across the residual stream for a range of values around $\Delta j = T_0 -1$ (s. Fig. \ref{Fig4}).
In the first layers of \textit{Gemma-2-2B}, we observe a pronounced maximum in both $\overline{C}{}^{(\ell)}_{\Delta}$ and $\overline{C}{}^{(\ell)}_{\varphi}$ at a value of $\Delta j = T_0$, signaling correlations between the same tokens of each subsequence.
We note that for $\overline{C}{}^{(\ell)}_{\Delta}$, this maximum is already present within the first layer of the model $(\ell=1)$ and therefore does not result from a composition of components across layers.
As we proceed through the residual stream, we find that the position of this maximum gradually shifts to $\Delta j = T_0 - 1$, predominately within a crossover region between $\ell\approx 30$ and $\ell\approx 40$.
Simultaneously, in this crossover region, $\overline{C}{}^{(\ell)}_{\vartheta}$ changes sign.
These results suggest that the induction behavior of \textit{Gemma-2-2B} is predominately composed within a localized subset of layers within the second half of the residual stream.

\begin{figure}
  \centering
  \includegraphics[width=\linewidth]{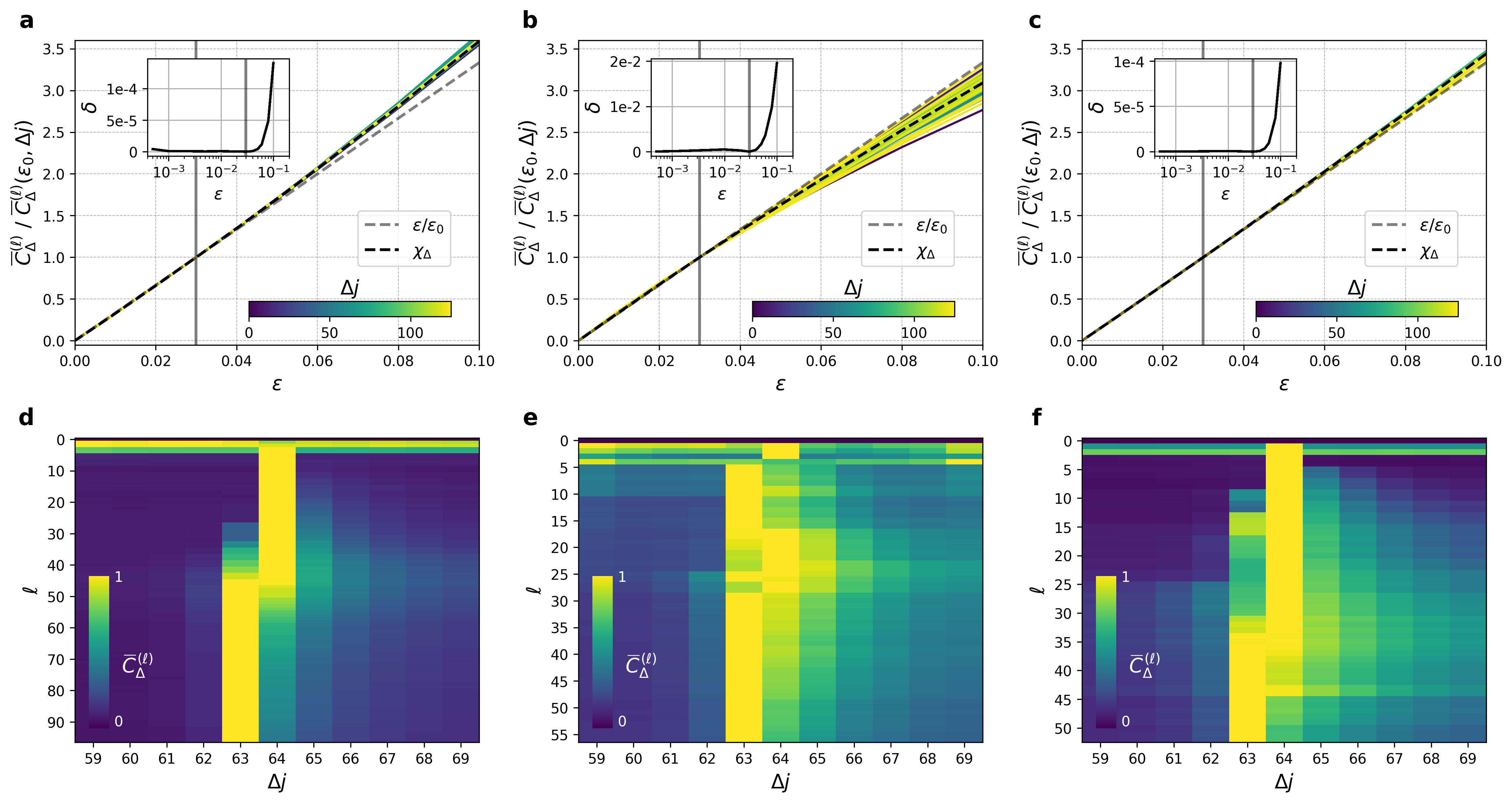}
  \caption{\textbf{Emergence of induction across models.} 
  \textbf{(a, b, c)} The ratios $\overline{C}{}^{(\ell)}_{\Delta} / \overline{C}{}^{(\ell)}_{\Delta}(\Delta j, \varepsilon_0)$ for the last layer $(\ell = 2L)$ in \textit{GPT-2-XL} (a, $L=48$), \textit{Llama-3.2-3B} (b, $L=28$), and \textit{Gemma-2-2B} (c, $L=26$), display the same universal scaling behavior for sufficiently weak perturbations $\varepsilon<\varepsilon_0$, where $\chi_{\Delta} \approx \varepsilon/\varepsilon_0$ (grey dashed lines), and $\varepsilon_0=0.03$ (grey vertical lines). Across all models, the relative deviation from the scaling function, $\delta=(\overline{C}{}^{(\ell)}_{\Delta} / \overline{C}{}^{(\ell)}_{\Delta}(\Delta j, \varepsilon_0) - \chi_\Delta) / \chi_\Delta$, averaged over all values of $\Delta j$, signals scale-invariant behavior over two orders of magnitude in $\varepsilon$ (insets). \textbf{(d, e, f)} Response functions $\overline{C}{}^{(\ell)}_{\Delta}$ for \textit{GPT-2-XL} (d), \textit{Llama-3.2-3B} (e), and \textit{Gemma-2-2B} (f), for varying values of $\Delta j$ and $\ell$. For each value of $\ell$, $\overline{C}{}^{(\ell)}_{\Delta}$ is normalized to its maximum value within the displayed range of $\Delta j = 59 - 69$. All data shown is obtained using the same data set as in Fig. \ref{Fig2}-\ref{Fig4}.}
  \label{Fig5}
\end{figure}

To understand to what extent our findings universally hold across models, we apply our method to two other models of similar size, \textit{GPT-2-XL} and \textit{Llama-3.2-3B}, using the same data set as described above (s. Fig. \ref{Fig5}).
Across all models, we observe the same universal scaling of $\overline{C}{}^{(2L)}_{\Delta}$ over two orders of magnitude in the scale parameter $(\varepsilon=5\cdot10^{-4} - 5\cdot10^{-2})$.
This result confirms the applicability of our method across models and provides evidence that the scale-invariant regime of the response is in fact a universal property across LLMs.

We extend our measurement of $\overline{C}{}^{(\ell)}_{\Delta}$ across the residual stream, and observe that both \textit{GPT-2-XL} and \textit{Llama-3.2-3B} display the characteristic shift of the maximum from $\Delta j = T_0$ to $\Delta j = T_0 - 1$, signaling the onset of induction behavior.
For \textit{GPT-2-XL}, this onset is characterized by a smooth transition, localized to positions $\ell\approx30-60$, while for \textit{Llama-3.2-3B}, there exist stronger fluctuations between adjacent layers and the maximum of the response shifts multiple times already in early layers before settling to its final value at $\Delta j = T_0 - 1$.
These results suggest qualitative differences between multi-layer models in forming induction behavior and serve as a benchmark for future circuit analyses within these models. 

\section{Conclusion}

In this work, we have observed the emergence of induction signatures within \textit{Gemma-2-2B}, \textit{GPT-2-XL} and \textit{Llama-3.2-3B} by probing the models' response to weak perturbations of the residual stream.
For each model, we found a pronounced universal regime, in which the response functions remain scale-invariant under changes in perturbation strength and thereby demonstrated the robustness of our method in quantifying the induction behavior.
Beyond the induction mechanism and models studied within this work, we expect our method to provide valuable information about the complex collective interplay of components within LLMs.

\subsection*{Limitations}

In this work, we focused solely on repeated sequences of random token to study induction behavior and, therefore, did not demonstrate the applicability of our method to real text sequences.
Furthermore, while our method can be readily extended to study also higher order correlations between tokens, we only considered two-token response functions within this work.
We expect such higher order correlations to be particularly relevant for actual text sequences containing more complex correlations between tokens.

Additionally, in our empirical work, we did not provide a theoretical foundation for the observed scale-invariant response. 
Therefore, it would be valuable to investigate how such remarkably robust, universal behavior can exist in the presence of the complex non-linear interplay of LLM components, and how it emerges during training. 

Finally, we did not demonstrate the universality of scale-invariance beyond the models \textit{Gemma-2-2B}, \textit{GPT-2-XL}, and \textit{Llama-3.2-3B}, studied within this work.
Therefore, it needs to be verified if the same scaling behavior and emergence of induction signatures can also be found in other models.
In particular, to demonstrate the applicability of our method to state-of-the-art models, the robustness of the scale-invariant regime to changes in model size needs to be verified. 

\section{Related Work}

\textbf{Universality} is a central research topic within mechanistic interpretability to understand if common features and circuits form across models and tasks \cite{ZoomInAn70:online}. 
For LLMs, universal behavior could be identified by finding common attention heads across models \cite{Incontex4:online,gould2023successor,wang2022interpretability}, the reuse of circuits for different tasks \cite{merullo2023circuit}, by stitching parts of different models together \cite{baek2024generalization}, or by analyzing the emergence of capabilities during training \cite{tigges2024llm}. 
At the same time, experiments focusing on modular addition tasks \cite{NEURIPS2023_56cbfbf4}, group composition \cite{chughtai2023toy}, and comparisons of neuron activations between different training instances of a model \cite{gurnee2024universal} showed mixed results for such universality claims.

\textbf{Induction} heads and circuits were discovered in toy models \cite{AMathema54:online} and later also analyzed in larger models using different approaches, including interventions on attention heads \cite{Incontex4:online,crosbie2024induction}. 
Studies also revealed more nuanced attention heads such as anti-induction heads \cite{Incontex4:online}, negative name mover heads \cite{wang2022interpretability}, successor heads \cite{gould2023successor}, copy suppression heads \cite{mcdougall2023copy}, semantic induction heads \cite{ren2024identifying}, and provided evidence for the emergence of induction during training \cite{singh2024needs}, and through the interplay of model components in toy models \cite{chen2024unveiling}.

Furthermore, our work uses weak perturbations to probe downstream token correlations in the residual stream, and therefore lies at the boundary between \textbf{intervention} and \textbf{observation} studies \cite{bereska2024mechanistic,rai2024practical}. 

In \textbf{observation} studies, layerwise interpretations of the model's output could be gained by applying the final classification layer \cite{interpre8:online}, and learned affine transformations \cite{belrose2023eliciting} to intermediate model layers or in between layers \cite{din2023jump}. Observation studies also used layerwise projections onto the space of vocabulary items \cite{dar2022analyzing}, cross-attention in encoder-decoder architectures \cite{langedijk2023decoderlens}, attention head specific transformations \cite{sakarvadia2023attention}, and analyzed the information of tokens about future tokens \cite{pal2023future}. Further post-hoc methods used attention scores \cite{abnar2020quantifying} and attribution scores and trees \cite{hao2021self} to uncover information flow in and between individual layers of transformers. 

\textbf{Intervention} studies enabled a study of the causal structure of LLMs \cite{Geiger2023CausalAA}. A large part of this work focused on activation patching methods such as causal tracing \cite{meng2022locating}, interchange intervention \cite{geiger2022inducing}, causal mediation \cite{vig2020investigating}, and causal ablation \cite{wang2022interpretability}. 
Activation patching methods could also reveal a robustness of the model to removing the majority of heads \cite{vashishth2019attention}, and removing and swapping adjacent layers \cite{lad2024remarkable}, to gain insights into the layerwise interference process, including how attention heads and MLP sublayers form representations \cite{geva2023dissecting}. 
Further extensions of such activation patching work include path patching which enabled a study of the effect of attention heads on the model's output \cite{goldowsky2023localizing}, and subspace activation patching \cite{geiger2024finding}. 
Furthermore, attribution patching methods used linear approximations to improve the scalability of activation patching \cite{Attribut16:online,syed2023attribution,ferrando2024information} and revealed the influence of single model components on the information flow within transformer models.

Finally, also relevant to our work is the method of Deep Taylor Decomposition \cite{montavon2017explaining}, which was used to extract relevancy maps of tokens for specific classification tasks \cite{chefer2021transformer}. 
In contrast, the correlation matrices (\ref{eq:CDelta}-\ref{eq:Ctheta}) used within our work can be interpreted to contain the relevancy \textit{between} tokens. 

\newpage 

\bibliographystyle{unsrtnat}
\bibliography{LLM_Response}

\appendix

\section{Additional computational details}

The experiments described within this work were performed using the TransformerLens library \cite{nanda_transformerlens_2022} which enables accessing and modifying activations within transformer models. All experiments described were performed on a single NVIDIA A100 GPU from a cloud provider. An overview of the pre-trained models used within this work can be found in Table \ref{tab:models}.

\begin{table}[h]
    \centering
    \caption{List of models used within this work.}
    \begin{tabular}{ccccc}
    \toprule
         Model name & Link & License & Citation \\
         \hline
         Gemma-2-2B       & \href{https://huggingface.co/google/gemma-2-2b}{HuggingFace: Gemma-2-2B}  & \href{https://ai.google.dev/gemma/terms}{Gemma Terms of Use} & \cite{gemma_2024} \\
         Llama-3.2-3B-Instruct     & \href{https://huggingface.co/meta-llama/Llama-3.2-3B-Instruct} {HuggingFace: Llama-3.2-3B-Instruct}  & \href{https://github.com/meta-llama/llama-models/blob/main/models/llama3_2/LICENSE} {Llama 3.2 Community License} & \cite{dubey2024llama}\\
         GPT-2-XL       & \href{https://huggingface.co/openai-community/gpt2-xl}{HuggingFace: GPT2} & \href{https://github.com/openai/gpt-2/blob/master/LICENSE}{Modified MIT License} & \cite{radford_language_2019} \\
    \bottomrule
    \end{tabular}
    \label{tab:models}
\end{table}

\end{document}